 \documentclass[10pt, conference, compsocconf]{IEEEtran}
\usepackage[utf8]{inputenc} 
\usepackage[T1]{fontenc} 
\usepackage{graphicx}
\usepackage{caption}
\usepackage[ruled,vlined]{algorithm2e}
\usepackage{subcaption}
\usepackage[hyphens]{url}
\usepackage{hyperref}
\usepackage{amsmath}
\usepackage{amssymb}
\usepackage{xcolor}

\ifCLASSINFOpdf
\else
\fi
\newcommand{\etal}{\textit{et~al.}}

\begin{document}
%
\title{RADDet: Range-Azimuth-Doppler based Radar Object Detection 
\\for Dynamic Road Users}

\author{\IEEEauthorblockN{Ao Zhang$^1$, Farzan Erlik Nowruzi$^{1, 2}$, Robert Laganiere$^{1, 2}$}
	\IEEEauthorblockA{
		\noindent\parbox{.27\textwidth}{\centering University of Ottawa$^1$}
		\noindent\parbox{.27\textwidth}{\centering Sensorcortek Inc$^2$}
		\\
		 \noindent\parbox{.27\textwidth}{\centering Ottawa, Ontario, Canada}
		 \noindent\parbox{.27\textwidth}{\centering Ottawa, Ontario, Canada}
		\\
		azhan085@uottawa.ca, fnowr010@uottawa.ca, laganier@eecs.uottawa.ca}
	

}


%


\maketitle

\begin{abstract}
	
Object detection using automotive radars has not been explored with deep learning models in comparison to the camera based approaches. This can be attributed to the lack of public radar datasets. In this paper, we collect a novel radar dataset that contains radar data in the form of Range-Azimuth-Doppler tensors along with the bounding boxes on the tensor for dynamic road users, category labels, and 2D bounding boxes on the Cartesian Bird-Eye-View range map. To build the dataset, we propose an instance-wise auto-annotation 
method. Furthermore, a novel Range-Azimuth-Doppler based multi-class object detection deep learning model is proposed. The algorithm is a one-stage anchor-based detector that generates both 3D bounding boxes and 2D bounding boxes on Range-Azimuth-Doppler and Cartesian domains, respectively. Our proposed algorithm achieves 56.3\% AP with IOU of 0.3 on 3D bounding box predictions, and 51.6\% with IOU of 0.5 on 2D bounding box prediction. Our dataset and the code can be found at \textcolor{blue}{\textit{\url{https://github.com/ZhangAoCanada/RADDet.git}}}.

\end{abstract}

\begin{IEEEkeywords}

Radar; Range; Azimuth; Doppler; Cartesian; Object Detection; Auto-annotation; Deep Learning

\end{IEEEkeywords}

%
\IEEEpeerreviewmaketitle

\section{Introduction}

Automotive radar, also known as Frequency Modulated Continuous Wave (FMCW) radar, has been widely adopted in Advanced Driver Assistance Systems (ADAS). Compared to other sensors such as camera and Lidar, radar is robust against adverse weather and lighting conditions, and is a less expensive~\cite{Ref:HighResolutionRadarDataset} sensor. The drawbacks of radar sensor are its low output resolution and high noise levels. Despite that, recent studies~\cite{Ref:ProbabilisticOriented, Ref:DeepRadarDetector, Ref:DeepOpenSpace} have shown the feasability of various tasks such as object detection and pose estimation using deep learning algorithms.

\begin{figure}[!t]
	\centering
	\includegraphics[width=3.3in]{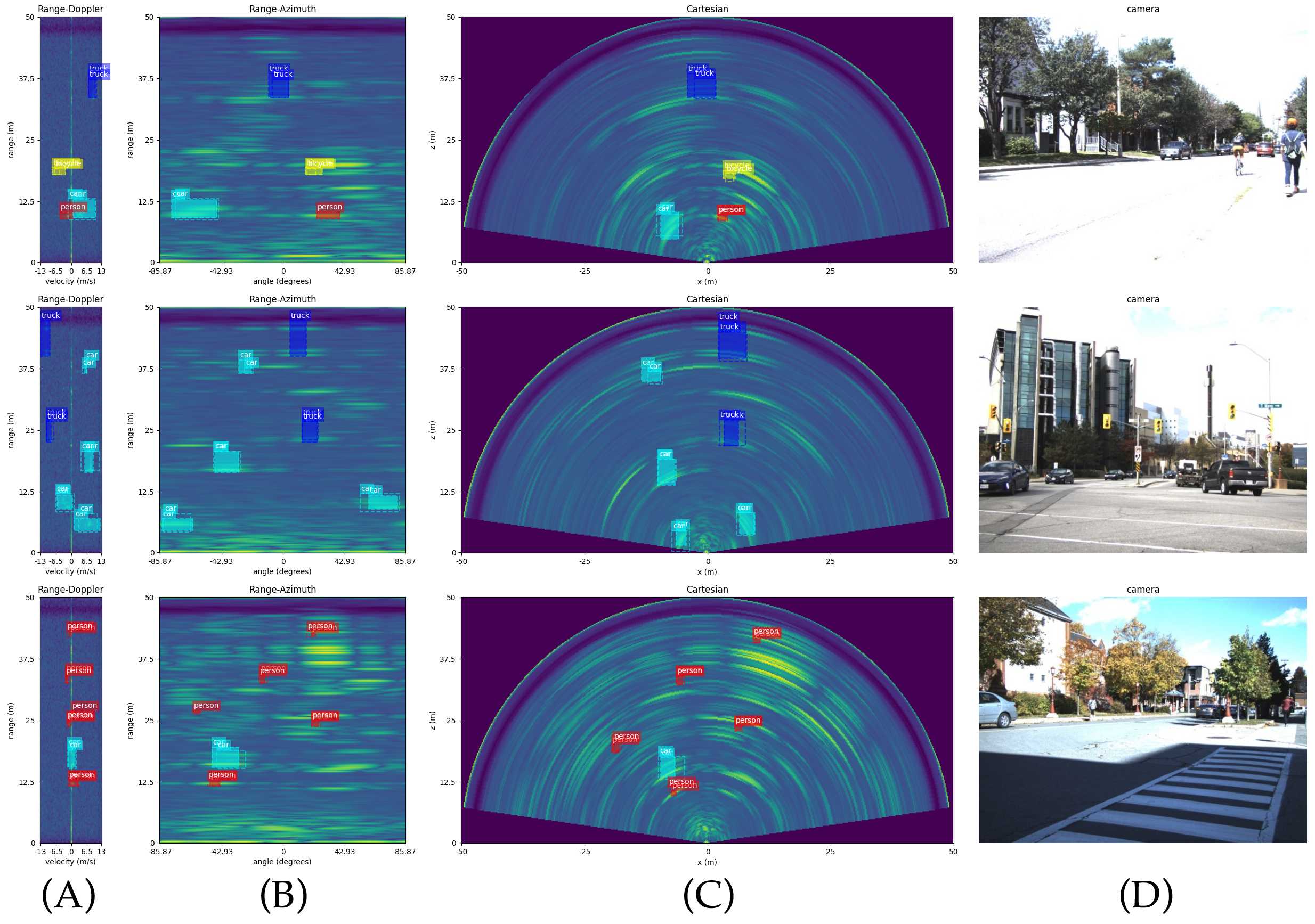}
	\caption{Instances from our dataset along with the predictions of our model. 3D boxes represented in Rang-Doppler spectrum (A) and Range-Azimuth spectrum (B). (C) The 2D representation of boxes in Cartesian coordinates. (D) The camra view used for visualization. Ground truth boxes are labelled with facecolors and predictions are without facecolors.}
	\label{F:DatasetSample}
\end{figure}

Raw data from FMCW radar, known as Analog-to-Digital (ADC) signals, are hardly readable for human observers. Thus, they are often passed into radar Digital Signal Processing (DSP) for interpretation. Sparse point cloud representation is the most popular output format for radar DSP due to its similarity to the output format from Lidar. Another widely used output format is a 3D tensor representation illustrating the information in range, angle and velocity. This format is often called Range-Azimuth-Doppler (RAD) spectrum. A sample is shown as the left two columns from Figure~\ref{F:DatasetSample}. RAD is the main representation that is used to extract the point-cloud representation. Since the Range-Azimuth dimensions are in the form of polar coordinates, they are often transformed into Cartesian coordinates for higher readability. There are two categories of approaches to extract RAD data from ADC signals, namely Fast Fourier Transform (FFT) and MUSIC~\cite{BG:MUSIC}. Although MUSIC provides higher precision, it is computationally expensive. In this research, we study the object detection with RAD spectrums achieved from applying FFT.

In recent years, deep learning based object detection algorithms have been widely explored in the image domain. In the radar domain, although object detection has gained a certain level of popularity, it is still an area that has not been explored to the full extent. This is due to several reasons. First, due to various input and output representations of the radar data, the focus of the research papers~\cite{Ref:ProbabilisticOriented, Ref:DeepRadarDetector, Ref:3DRadarCube, Ref:ClusterSegmentation} are split between those. Second, there is shortage of public datasets that provide the raw input representations and capacity to benchmark the studies in this field. Thus, the researchers chose to build their own dataset. 
One common observation that can be found in most radar research projects is that they only target dynamic objects~\cite{Ref:ProbabilisticOriented, Ref:3DRadarCube}, since the detections of dynamic objects are richer than static ones.

In this paper, we introduce a novel publicly available dataset and propose a new object detection model for dynamic road users. Our contributions are as follows:

\begin{itemize}
	\item A novel dataset that contains radar data in the form of RAD representations with the corresponding annotations for various object classes is introduced. The dataset is available\footnote{\label{webpage}\url{http://www.site.uottawa.ca/research/viva/projects/raddet/index.html}} for future research in this field. 
	\item An automatic annotation method that generates ground-truth labels on all dimensions of RAD data and in the Cartesian form is proposed. 	
	\item A new radar object detection model is proposed. Our model employs a backbone based on ResNet~\cite{BG:ResNet}. The ultimate form of the backbone is achieved after systematic explorations of deep learning models on the radar data. Inspired by YOLO head~\cite{Ref:YOLOv4}, we propose a new dual detection head, with a 3D detection head on RAD data and a 2D detection head on the data in Cartesian coordinates.	
	\item Our proposed model is extensively evaluated against the well known image based object detection counter-parts. Results show that our proposal achieves the state of the art performance in the object detection task with radar data. The significance of our results shows that radar is a viable competitor to camera and lidar based object detection methods.
\end{itemize}

%
%
%
%
%
%

\section{Related Work}

One common issue of autonomous radar research is the lack of proper public datasets. This is due to the radar hardware variance and the highly configurable property of the radar itself. For example, a different number of on-board transmitters and receivers will result in different data resolutions. Also, with different frequency sweep magnitudes and sampling rates, the maximum detection range and range resolution will behaviour totally different. Thus, most radar researchers chose to create the dataset themselves using sensor fusion techniques.


In the previous work \cite{Ref:DeepRadarDetector, Ref:RadarCamFusion, Ref:DeepOpenSpace, Ref:3DRadarCube, Ref:RADVehicle, Ref:ProbabilisticOriented}, cameras and Lidars are used along with the radar for building their datasets.
In order to improve the quality of the dataset, some researchers~\cite{Ref:HighResolutionRadarDataset} even use multiple high-precision sensors for annotating radar data, namely Lidar, Synthetic Aperture Radar (SAR) and cameras.

Except for employing different sensors, another big difference among all the datasets mentioned above is the input radar data format. Generally, it can be classified into two categories: cluster-based radar input format and spectrum-based radar input format.

Cluster-based input format is the data format that contains a cluster of points with each point illustrating the location of the object. To generate such format, some traditional cluster-based pre-processing methods need to be applied. Schumann~\etal~\cite{Ref:ClusterLSTM} use DBSCAN~\cite{BG:DBSCAN} to differentiate and extract all instances in radar frames, then annotate each instance with the corresponding object in the camera frames. 
One challenge of the cluster-based methods is that data could be mistakenly merged or separated due to the limited range resolution and the high noise level. Palffy~\etal~\cite{Ref:3DRadarCube} manage to overcome such problem by adding cropped spectrum-based data to the clusters and employ 3D convolutions for processing the input radar cube.

Spectrum-based input format such as RAD spectrum
keeps the consistency of the input data without any discretization. This will result in a performance boost for most deep learning algorithms. However, it also increases the complexity of the input data, which could potentially impact the runtime speed of those algorithms. Brodeski~\etal~\cite{Ref:DeepRadarDetector} successfully conduct 3D point-cloud segmentation and pose estimation using Range-Doppler-Angles spectrum, where angles stand for a combination of azimuth and elevation angles. Nowruzi~\etal~\cite{Ref:DeepOpenSpace} use Range-Azimuth-Doppler spectrums as inputs for free space segmentation. Major~\etal~\cite{Ref:RADVehicle} sperate Range-Azimuth-Doppler inputs along $3$ axes and feed them into the networks before integrating them together. 
The Range-Azimuth spectrum is used in \cite{Ref:RadarCamFusion, Ref:ClassificationRadarSpectra, Ref:ProbabilisticOriented} for object detection and classification. 

Based on different types of input, various base models from computer vision domain were used with the Radar data. For cluster-based input,
Schumann~\etal~\cite{Ref:ClusterSegmentation} explore multi-scale grouping module (MSG) from PointNet++~\cite{Ref:PointNet++} for point-cloud based segmentation. For spectrum-based input, VGG~\cite{BG:VGG} is chosen as the base model in \cite{Ref:ClassificationRadarSpectra}. Brodeski~\etal~\cite{Ref:DeepRadarDetector} build the high-level architecture based on Faster-RCNN~\cite{Ref:FasterRCNN}. U-Net~\cite{Ref:U-Net} and FCN~\cite{Ref:FCN} architectures are the most popular ones among recent radar research projects~\cite{Ref:ProbabilisticOriented, Ref:RadarGhostObject, Ref:DeepOpenSpace}.

For object detection, the detection head is one of the most researched topics. Typical anchor-based one-stage detection heads~\cite{Ref:SSD, Ref:YOLOv4, Ref:FocalLoss} are still leading in this field. 
On the other hand, anchor-free one-stage detection heads~\cite{Ref:FCOS} have gained certain popularity recently and the performance is gradually catching up with anchor-based detection heads. Self-attention layers~\cite{Ref:SAGAN, Ref:SAUnet} show a growing potential in achieving further improvements in this task. Two-stage detection heads~\cite{Ref:FasterRCNN} are always considered to be able to achieve higher performance but much slower than one-stage heads. In radar research, \cite{Ref:ProbabilisticOriented} proposes a probabilistic oriented object detection head for oriented 2D box detection.

\begin{figure*}[!t]
	\centering
	\includegraphics[width=.99\textwidth]{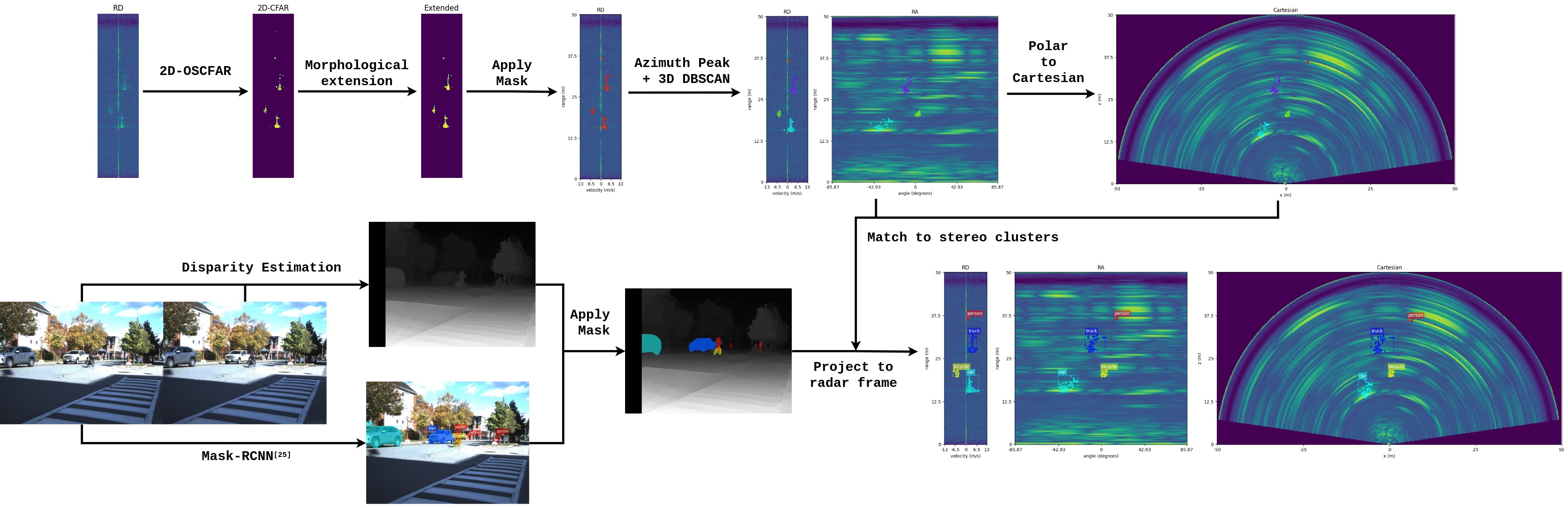}
	\caption{The proposed auto-annotation method for the ground truth generation that includes an enhanced radar pre-processing and stereo depth estimation.}
	\label{F:InstancizeProposal}
\end{figure*}

Compared with the previous research projects, our proposed dataset takes full-size information of radar data, including velocity, range, angle and the corresponding position in Cartesian coordinates. In addition, we choose ResNet~\cite{BG:ResNet} and YOLO~\cite{Ref:YOLOv4} as the base models for our proposed deep learning algorithm. In return, it boosts the performance of object detection on radar data with the optimal run time speed.

\section{Dataset Preparation}

In this section, we introduce the hardware setups of our radar and stereo cameras. We propose a new auto-annotation method that tackles the challenging radar annotation task. Extracted annnotations are then manually validated to ensure the quality of dataset. Finally, we report the statistical analysis of our dataset.


\subsection{Hardware Setup}
\label{Section:ProjectionMatrix}

{The sensors used in data collection consist of a \textit{Texas Instruments} AWR1843-BOOST\footnote{\url{www.ti.com}} radar and a pair of DFK 33UX273\footnote{\url{www.theimagingsource.com}} stereo cameras from \textit{The Imaging Source}. }
The sensors calibration 
is implemented with a self-made trihedral corner reflector 
under the instruction of \cite{BG:Calibration}. 
{Random locations were set for the calibration target with distances to the sensors ranging from $5 m$ to $50 m$. The projection matrix from the 3D stereo frame to the bird-eye-view radar frame is computed with least square optimization.} Raw ADC data captured from the radar have a shape of $(256, 8, 64)$.

\subsection{Annotation Method}



Traditional DSP for FMCW radar is divided into two steps. First, FFT is performed on each dimension of received ADC signals. The primary output of this step is RAD spectrum. 
In our research, zero-padding is used while computing FFT on Azimuth dimension and the RAD outputs are generated with a shape of $(256, 256, 64)$. 
Second, Constant False Alarm Rate (CFAR) is employed for filtering out the noise signals on Range-Doppler dimensions. There are two major CFAR algorithms, namely Cell-Averaging CFAR (CA-CFAR) and Ordered-Statistic CFAR (OS-CFAR). OS-CFAR is normally more preferable for academic usages due to its high-quality outputs, while CA-CFAR is often used in the industry because of the speed. 
The output of this step is often transformed to Cartesian coordinates and presented as a point-cloud., which is the base of  cluster-based radar data analysis in various applications.


By analyzing the output of traditional radar DSP, we found that the detection rate of dynamic road users is significantly higher than static road users. This is due to the fact that dynamic road users usually have richer information on the Doppler axis. Therefore, we set our targets to dynamic road users only. On the other hand, traditional OS-CFAR can generate false detections in some cases even with fine-tuned parameters. For example, it could ignore points when the objects are too big, or it could mislabel noise points as detections. Our proposed annotation method can effectively solve these issues.

On the Range-Doppler (RD) outputs from 2D OS-CFAR, rigid bodies such as vehicles can be easily detected due to the speed coherence on the Doppler axis. For pedestrians, different motions of different body parts may result in various output patterns on the RD spectrum~\cite{Ref:DeepRadarDetector}. However, when the radar's range resolution reaches a certain level, the complexity of human bodies' motion can hardly be observed. Thus, they also can be considered as rigid bodies. An example is shown in Figure~\ref{F:InstancizeProposal}. One property of the rigid bodies appearing on the RD spectrum is that they are often presented as linear patterns, despite the angle differences between the objects and the radar. Thus, by connecting the discrete patterns on the RD spectrum, we successfully enrich the detection rate of traditional 2D OS-CFAR.

However, this method provides poor detections when objects are at the same range with a similar speed. DBSCAN~\cite{BG:DBSCAN} can effectively alleviate such problems. This way, we can precisely extract object instances from RAD data. 
The dataflow of the proposed radar pre-processing method is shown in Figure~\ref{F:InstancizeProposal}.

With our proposal, radar instances can be easily extracted from RAD spectrum. 
However, classifying radar-only instances is a challenging task. In this research, we rely on stereo vision 
for the category labelling. The whole process is described as follow. 
First, disparity maps are generated from the rectified stereo image pairs using Semi-Global Block Matching algorithm with OpenCV\footnote{https://opencv.org/}. 
Then, the pre-trained Mask-RCNN~\cite{Ref:MRCNN} model is applied on the left images to extract instance-wise segmentation masks. The prediction masks are then projected onto the disparity maps. Finally, using the triangulation, the instance-wise point cloud outputs with predicted categories are generated. Afterwards, the point cloud instances are transformed to the radar frame using the projection matrix obtained in \ref{Section:ProjectionMatrix}.

Our auto-annotation method is developed with the obtained radar and stereo instances. Figure~\ref{F:InstancizeProposal} shows the whole pipeline of the process. 
Using this method, we can retrieve annotations for around $75\%$ of all the objects in the dataset. 
Multiple factors can contribute to the errors in the annotations, such as errors in radar instance extraction due to the sensor noise, errors in depth estimation from the stereo cameras, and errors in predictions of the Mask-RCNN~\cite{Ref:MRCNN}.  
In addition, since the Field of View (FOV) of the stereo cameras is considerably smaller than the radar's, a certain number of objects are left unlabelled. To solve these issues, manual corrections are conducted 
after the auto-annotation process to generate the dataset. 

\subsection{Dataset Analysis}

\begin{figure}[h]
	\centering
	\begin{subfigure}{0.25\textwidth}
		\centering
		\includegraphics[width=.99\linewidth]{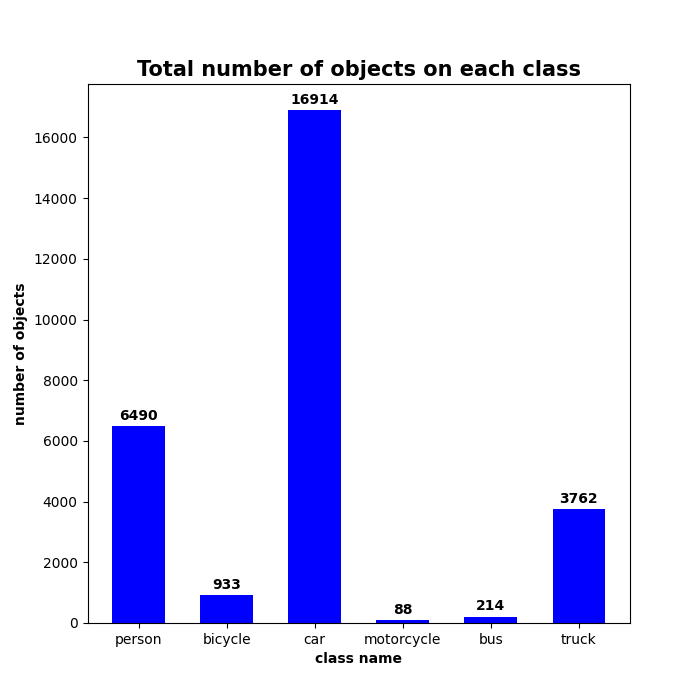}
	\end{subfigure}%
	\begin{subfigure}{0.25\textwidth}
		\centering
		\includegraphics[width=.99\linewidth]{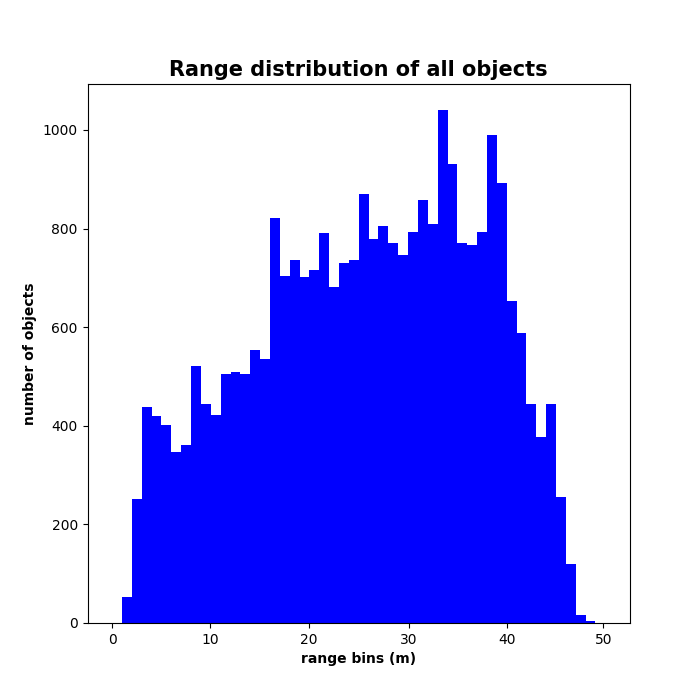}
	\end{subfigure}
	\caption{Left: distribution of the number of objects over all categories. Right: range distribution over all objects.}
	\label{F:DatasetAnalysis}
\end{figure}

The data capture sessions were conducted in sunny weather conditions at several random locations during the time span from September to October. The sensors were set up on the sidewalks and facing the main roads, as shown in Figure~\ref{F:DatasetSample}. 
After removing the corrupted frames, a total number of $10158$ frames are collected to build our dataset.
As for the statistical analysis of the dataset, Figure~\ref{F:DatasetAnalysis} shows the details.

We split our dataset into train and test sets with the ratios of $80\%$ and $20\%$ respectively using class-wise random sampling. This ensures that both our train and test sets are following the overall distributions shown in Figure~\ref{F:DatasetAnalysis}. Furthermore, we split the train set to $90\%$ for training and $10\%$ for validation during the experiments to find the optimal coefficients and models.

\section{RADDet}

\begin{figure*}[!t]
	\centering
	\includegraphics[width=.90\textwidth]{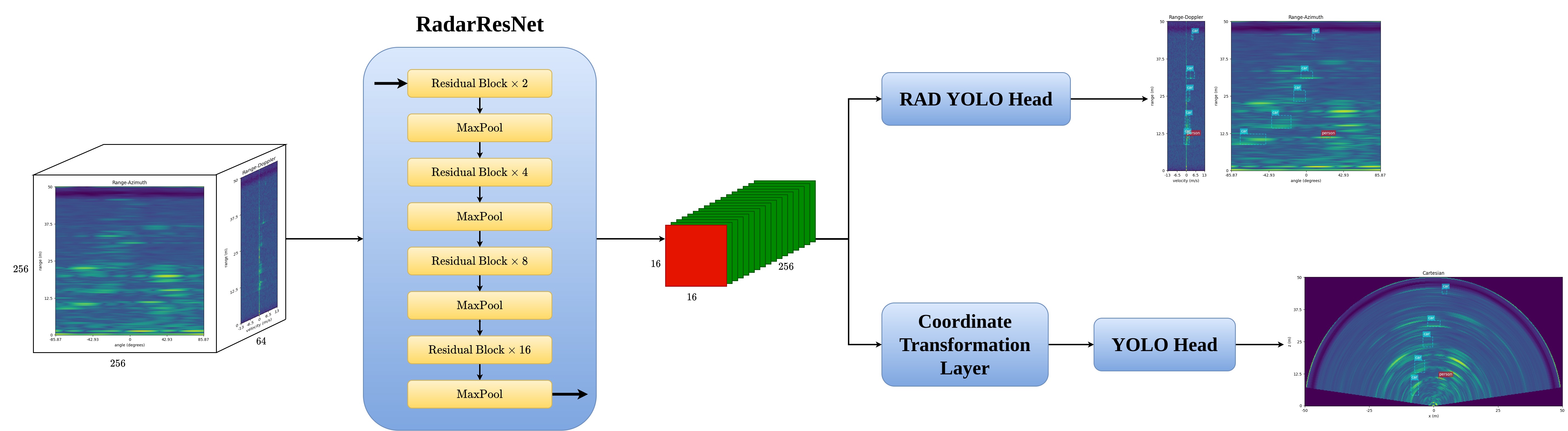}
	\caption{Dataflow of our proposed networks architecture.}
	\label{F:Architecture}
\end{figure*}

The state-of-the-art image-based object detection algorithms consist of $3$ parts, a backbone, a neck and a detection head~\cite{Ref:YOLOv4, Ref:FCOS, Ref:FocalLoss}. Inspired by that we formulate our backbone networks based on widely used ResNet~\cite{BG:ResNet}. In the image domain, the neck layers are used to extract ouputs in multiple levels in order to handle the scale variations of the objects. However, unlike images, where the distance changes the size of the objects due to pespective geometry, radars reveal the true scale of the objects. Therefore, multi-resolution neck layers are not considered in our research. Finally, we propose a novel dual detection head based on the well-known anchor-based algorithm YOLO~\cite{Ref:YOLOv4}. Figure~\ref{F:Architecture} shows the dataflow of our proposed architecture. Details are introduced in the following sections.

\subsection{Global Normalization of the Input}

The raw input to our model is a Range-Azimuth-Doppler tensor of size $(256, 256, 64)$, represented in the complex number form. As done in traditional radar DSP, we first extract the magnitude of the raw input and take log of it. In order to normalize the input, the mean value $V_{mean}$ and the variance value $V_{variance}$ are searched over the entire dataset $D_{dataset}$. Then, each input tensor is normalized as Equation~\ref{Eq:Normalization}.
\begin{equation}
	{I_i}_{norm} = \frac{(I_{i} - V_{mean})}{V_{variance}},\,\, I_{i} \in D_{dataset}
	\label{Eq:Normalization}
\end{equation}
where, ${I_i}_{norm}$ is the input to our backbone in the next step.

\subsection{RadarResNet}

The input, as introduced above, is quite different from a colored image that is used in image based methods. In order to adapt 2D convolutional layers to our task, we set Range-Azimuth axes as the input dimensions and the Doppler axis as the original channel size. Channel expansion and feature map downsampling methods are considered during the architecture search. After a large set of architecture exploration and experiments, we finalized our backbone network based on ResNet~\cite{BG:ResNet} and named it RadarResNet. Figure~\ref{F:Architecture} shows the RadarResNet that consists of two types of blocks, namely the residual blocks and the downsampling blocks. The residual blocks are of the same structure as the basic residual blocks in ResNet and the downsample blocks consist of one max-pooling layer. Activation functions are set as Rectified Linear Unit and Batch Normalization~\cite{BG:BN} is applied after each convolutional layer in the residual blocks. The final output size of the proposed backbone is $(16, 16, 256)$.

\subsection{Dual Detection Head}

Our detection head consists of two branches: a 3D and a 2D detection head. Both of them are anchor-based methods and with similar structures as YOLO~\cite{Ref:YOLOv4}. In our research, $6$ anchor boxes are defined for both 3D and 2D detection head branches
by 
using K-means clustering on all the ground truth boxes with the error rate threshold set to $10\%$. 


\subsubsection{3D Detection Head}

Since the outputs of our 3D detection head are 3D bounding boxes, some modifications are applied to the traditional YOLO Head in order to fit our task. First, the 
output size of the third 
dimension, Doppler dimension, is calculated with the same 
stride as other dimensions. In our case, the output size of the Doppler axis is $4$. Second, instead of convolving the feature maps into $(16, 16, num\_of\_anchors \times (5 + num\_of\_classes))$~\cite{Ref:YOLOv4}, the 3D head processes the feature maps into $(16, 16, 4 \times num\_of\_anchors \times (7 + num\_of\_classes))$, where $7$ stands for the objectness and the 3D box information. The 3D box information consists of the 3D center point [$x$, $y$, $z$], and the size [$w$, $h$, $d$]. Finally, the raw detection output is reshaped into the final output format $(16, 16, 4, num\_of\_anchors, 7 + num\_of\_classes)$, before it is fed into Non-maximum Suppression (NMS). Other operations such as box location calculation and interpretation, are structured in the same form as YOLO~\cite{Ref:YOLOv4}. For convenience, we named our 3D detection head as RAD YOLO Head.

\subsubsection{2D Detection Head}

\begin{figure}[!t]
	\centering
	\includegraphics[width=3.3in]{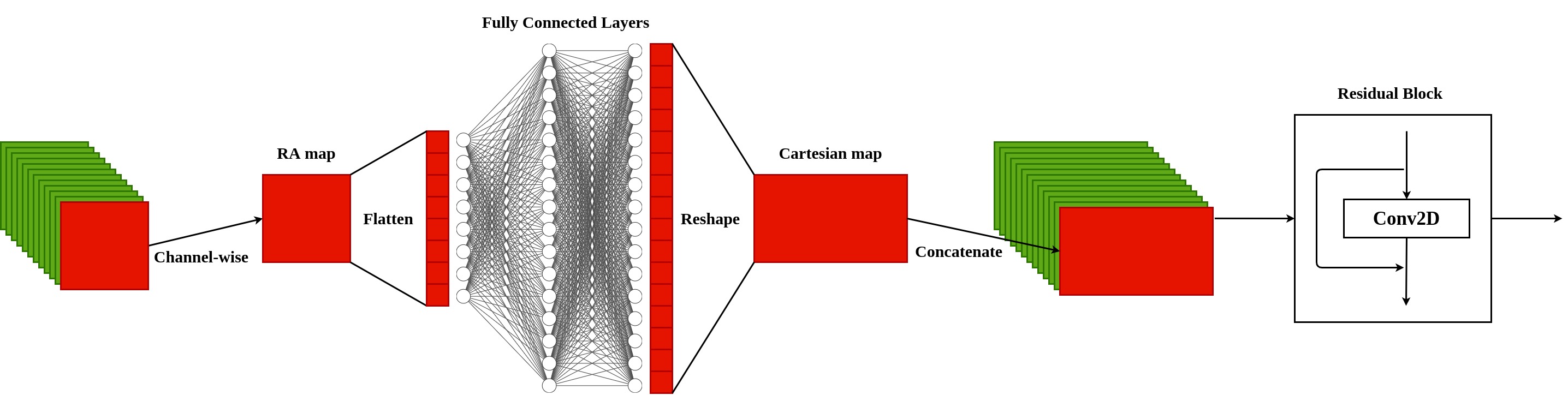}
	\caption{Coordinate Transformation Layer consists of Channel-wise fully connected layers acting as non-linear transformation for each feature map and a residual block to summarize the features.}
	\label{F:CoordinateTransformation}
\end{figure}

The 2D detection head consists of two parts; a coordinate transformation layer that transforms the feature maps from polar representation to the Cartesian form, and a classic YOLO Head~\cite{Ref:YOLOv4}. The structure of the coordinate transformation layer is shown in Figure ~\ref{F:CoordinateTransformation}.

Traditionally, the coordinate transformation from range and azimuth domain $[r, \theta]$ to cartesian width and depth domain represented by $[x, z]$, is formulated by the following equations.
\begin{equation}
\begin{split}
	x &= r \cdot \text{cos}(\theta) \\
	z &= r \cdot \text{sin}(\theta) \\
	\theta &\in [-\pi/2, \pi/2]
\end{split}
\label{eq:Pol2Cart}
\end{equation}
This non-linear transformation can double the size of the input along the width dimension in cartesian form. 

Our Coordinate Transformation Layer was inspired by the traditional method. The input feature maps are in the form $[r, \theta]$ with the size of $(16, 16, 256)$. They can be interpreted as $256$ RA feature maps with the shape of $(16, 16)$. Each of the RA feature maps is fed individually into two fully connected layers for non-linear transformation. The outputs of these features are then reshaped to $(32, 16)$ and concatenated to build the Cartesian feature maps of size $(32, 16, 256)$. Finally, since they are low-level feature maps, one residual block, with the same structure as the basic residual block in ResNet~\cite{BG:ResNet}, is structured for post-processing. The transformed outputs are then directly fed into YOLO Head for 2D box predictions.

\subsection{Loss Functions}

The outputs of the YOLO Head are normally split into $3$ tasks, namely box regression, objectness prediction and classification \cite{Ref:YOLOv4}. Various loss functions have been explored for these $3$ tasks to improve the performance. In our research, The total loss $L_{total}$ is formulated as Equation~\ref{Eq1:TotalLossFunction}.

\begin{equation}
	L_{total} = \beta \cdot L_{box} + L_{obj} + L_{class}
	\label{Eq1:TotalLossFunction}
\end{equation}

The loss function for box regression loss $L_{box}$ is chosen from \cite{Ref:YOLOv1}. Focal Loss~\cite{Ref:FocalLoss} is applied for training the objectness prediction loss $L_{obj}$. However, in order to remove the dominance of non-object predictions, we set the $\alpha$ coefficient to $0.01$ for negative samples in the Focal Loss. For the classification loss $L_{class}$, we use Cross Entropy~\cite{BG:CrossEntropy} as the loss function. During the training, we found that the box loss can easily dominate the total loss values. Therefore, we set another coefficient $\beta = 0.1$ for the box loss.

\section{Experiments}

We follow the same experimental setup of image based object detection literature for our model exploration and experiments. First, we explore several popular backbones, namely VGG~\cite{BG:VGG} and ResNet~\cite{BG:ResNet}. Next, usage of layers such as self-attention and channel-wise Multilayer Perceptrons (MLP) prior to extracting bounding boxes are explored. Among the various self-attention proposals from the literature, we selected two, SAGAN~\cite{Ref:SAGAN} and SAUNet~\cite{Ref:SAUnet}, for our experiments. Since our detection heads share the same backbone, we trained the backbones along with the RAD YOLO head as it is more complicated than the 2D detection head. Before training the 2D detection head, we froze the backbone layers after fine-tuned on the 3D detection head. The following sections show the comparisons of different backbones on both 3D detection head and 2D detection.

Hyper-parameters used in our experiments are defined as follows: batch size is set to $3$; initial learning rate is $0.0001$; the learning rate decay is $0.96$ with every $10$K steps after $60$K warm-up steps; the objectness threshold is $0.5$; the NMS thresholds for 3D and 2D detection heads are $0.1$ and $0.3$ respectively; the optimizer is set to Adam~\cite{BG:Adam} and the mean Average Precision is chosen as the metrics. All experiments are conducted with an RTX 2080Ti GPU and TensorFlow API\footnote{\url{https://www.tensorflow.org/}}.

\begin{table*}[!t]
	\begin{center}
		\begin{tabular}{ c | c c c c c c c c }
			\hline
			\\[-1ex]
			\textbf{Backbone Name} & \textbf{Detection Head} & \textbf{$\text{AP}_{0.1}$} & \textbf{$\text{AP}_{0.3}$} & \textbf{$\text{AP}_{0.5}$} & \textbf{$\text{AP}_{0.7}$} & \begin{tabular}{@{}c@{}} \textbf{Backbone} \\ \textbf{Params} \end{tabular} &  \begin{tabular}{@{}c@{}}\textbf{Head} \\ \textbf{Params} \end{tabular} & \begin{tabular}{@{}c@{}}\textbf{Inference} \\ \textbf{Time}($\tt ms$)\end{tabular} \\
			\\[-1ex]
			\hline
			\\[-1ex]
			VGG  & RAD YOLO Head & 0.685 & 0.499 & 0.209 & 0.035 & \textbf{2.10M} & 1.34M &  \textbf{10.5} \\
			{RadarResNet {(strided convolution)}}  & RAD YOLO Head & 0.727 & 0.538 & 0.243 & 0.054 & 6.81M & 1.34M & 75.3 \\
			RadarResNet {(max-pool)} & RAD YOLO Head & \textbf{0.764} & \textbf{0.563} & \textbf{0.251} & \textbf{0.059} & 6.73M & 1.34M & 75.2 \\
			RadarResNet+SA (SAGAN~\cite{Ref:SAGAN}) & RAD YOLO Head & 0.738 & 0.498 & 0.219 & 0.044 & 4.11M & 1.34M & 50.9 \\
			RadarResNet+SA (SAUNet~\cite{Ref:SAUnet}) & RAD YOLO Head & 0.752 & 0.559 & \textbf{0.251} & 0.053 & 7.55M & 1.34M & 78.7 \\
			RadarResNet+MLP & RAD YOLO Head & 0.759  & 0.521 & 0.217 & 0.043 & 6.86M & 1.34M & 77.6\\
			\\[-1ex]
			\hline
			\\[-1ex]
			VGG   & 2D YOLO Head & 0.757 & 0.685 & 0.484 & 0.189 & \textbf{2.10M} & 2.86M & \textbf{12.8} \\
			{RadarResNet {(strided convolution)}}   & 2D YOLO Head & 0.770 & 0.699 & 0.502 & 0.198 & 6.81M & 2.86M & 76.7 \\
			RadarResNet {(max-pool)} & 2D YOLO Head & 0.796 & 0.727 & 0.516 & \textbf{0.205} & 6.73M & 2.86M & 76.8 \\
			RadarResNet+SA (SAGAN~\cite{Ref:SAGAN}) & 2D YOLO Head & 0.787 & 0.686 & 0.452 & 0.145 & 4.11M & 2.86M & 52.6 \\
			RadarResNet+SA (SAUNet~\cite{Ref:SAUnet}) & 2D YOLO Head & \textbf{0.801} & \textbf{0.730} & \textbf{0.530} & 0.202 & 7.55M & 2.86M & 79.1 \\[1ex]
			\hline
		\end{tabular}
	\end{center}
	\caption{Average Precisions of different backbones on both detection heads.}
	\label{T3:BackboneAP}
\end{table*}


\subsection{Backbone Model Search}

Before training on the entire train set, we set up a framework with a small portion of the training data to find a proper output size of our backbone. After several trials on both VGG~\cite{BG:VGG} and ResNet~\cite{BG:ResNet} with different output sizes such as $(8, 8)$, $(16, 16)$ and $(32, 32)$. We found that using size $(16, 16)$ provides us with the highest performance in the 3D detection task. Therefore, we first fixed the output size of our backbone as $(16, 16, 256)$. Based on that, The total number of channel expansion blocks and downsampling blocks are calculated as $2$ and $4$. To adapt the image-based backbones to such structure, we modified each of them with proper channel expansion rates and downsampling rates. In VGG, we set the channel expansion rates of the first three blocks to $1$ and removed the pooling layer in the first block.

During the unit tests, we also found that the general performance of ResNet based backbones is higher than VGG. In addition, using max-pooling layers other than residual blocks for downsampling can significantly improve the performance as well. We then formulated our RadarResNet as illustrated in Figure~\ref{F:Architecture}. However, for ResNet-like architectures, the repeat times of the residual blocks vary when it comes to different tasks. Thus, we trained our RadarResNet with various sizes on the entire train set to find the optimal model. Table~\ref{T4:BackboneSearch} shows some examples of the backbone size exploration process. We finalized our RadarResNet with repeat times $2$, $4$, $8$, $16$ before each max-pooling layer. 

\begin{table}[!h]
	\begin{center}
		\begin{tabular}{ c | c c c c }
			\hline
			\\[-1ex]
			\textbf{Backbone Name} & \textbf{$\text{AP}_{0.1}$} & \textbf{$\text{AP}_{0.3}$} & \textbf{$\text{AP}_{0.5}$} & \begin{tabular}{@{}c@{}} \textbf{Backbone} \\ \textbf{Params} \end{tabular}\\
			\\[-1ex]
			\hline
			\\[-1ex]
			RadarResNet [1, 2, 4, 8] & \textbf{0.766} & 0.554 & 0.238 & 3.68M \\
			RadarResNet [2, 4, 4, 8] & 0.752 & 0.555 & 0.247 & 3.91M \\
			RadarResNet [4, 8, 8, 8] & 0.753 & 0.557 & 0.247 & 4.70M \\
			RadarResNet [2, 4, 8, 16] & 0.764 & \textbf{0.563} & \textbf{0.251} & 6.73M \\
			RadarResNet [4, 8, 8, 16] & \textbf{0.766} & \textbf{0.563} & 0.247 & 7.21M \\[1ex]
			\hline
		\end{tabular}
	\end{center}
	\caption{Backbone size searching samples with numbers inside ``[]'' indicating the repeat times of residual blocks before downsampling.}
	\label{T4:BackboneSearch}
\end{table}

\subsection{Detection Head Comparisons}

We conduct systematic comparisons between different backbones on both 3D and 2D detection heads in our experiments. Apart from the mentioned models, self-attention layers~\cite{Ref:SAGAN, Ref:SAUnet} are also explored. Our experiments show that self-attention layers convey the potential to dominate in object detection. Although the implementation of self-attention layers varies among different studies, the mechanism remains the same. In order to study the general impact of self-attention layers on our model, we employed the SAGAN~\cite{Ref:SAGAN} and SAUNet~\cite{Ref:SAUnet} individually on our RadarResNet. For convenience, we named these two backbones as RadarResNet+SA (SAGAN) and RadarResNet+SA (SAUNet). In addition, we experimented the channel-wise Multilayer Perceptrons (MLP) at the end of RadarResNet to form another backbone named RadarResNet+MLP. Table~\ref{T3:BackboneAP} shows the performance of the different backbones on different detection heads. The inference time is calculated by averaging the run time over the entire test set with batch size set to $1$.

For the RAD YOLO Head, the RadarResNet outperforms all other backbones due to several potential reasons. First, similar to image based models, the residual blocks outperform sequential convolutional layers in radar object detection as the original input signals are amplified during the process. Second, both types of self-attention layers are designed for images and are not able to boost performance in radar-based tasks. Table~\ref{T3:BackboneAP} also shows that max-pooling layers are more preferable than the strided convolutional layers in our task. 
As the 2D detection head already contains MLP, 
we excluded the RadarResNet+MLP from 2D detection experiments.

For the 2D YOLO Head, although RadarResNet+SA (SAUNet) indicates slightly better performance, the RadarResNet shows better overall precision. In addition, the inference speed of RadarResNet is considerably higher than the other one. This further proves that image-based self-attention layers may not be suitable for radar-based tasks.

\subsection{Discussion}

 Some sample results on the test set are shown in Figure~\ref{F:DatasetSample}. There are two common false detections found during our experiments, mis-classification and false positive detection. We believe both problems can be a resulting from the dataset. Although we have created and labeled a large dataset, the class labels of our dataset suffer from label imbalance shown in Figure~\ref{F:DatasetAnalysis}. In addition, the limited dataset size constrains the model to underperform and more performance will be achieavable with an expansion on the dataset size. Some ambiguities of the classes can also be found resulting from the sizes of the objects in the dataset. For example, assigning mini-vans to the ``truck'' class and assigning large SUVs to the ``car'' class confuses the model due to the fact that the signal traces of these objects look similar in radar frames.



\section{Conclusion}

In this paper, we introduce a comprehensive publicly available dataset with raw radar observations for multi-class object detection of dynamic road users. Details of Range-Azimuth-Doppler (RAD) based auto-annotation method for generating annotated radar instances is provided. Furthermore, we propose a robust and fast deep learning model for RAD based object detection and benchmark it against the other well known models. To the best of our knowledge, this is the first model that inputs raw radar inputs and predicts the position, speed and category of the dynamic road users. Despite the large size of our proposed dataset, it is still much smaller than its image based counterparts. Collection of more targeted data can balance the object categories and result in better performance of the models. Our proposed model achieves fast inference speeds on an RTX 2080Ti. However, deploying it on embedded platforms will result in a drastic slow down. To address this, a hardware aware model exploration needs to be conducted. Lastly, we believe using the self-attention layers to target this specific data rather than employing them from image based literature could significantly enhance performance of such models.

We sincerely hope that this research and the dataset can bridge the gap between image or Lidar based object detection and radar-based object detection, and inspire the development of more algorithms on automotive radars.



\section*{Acknowledgment}

The authors would like to thank SensorCortek\footnote{\url{https://sensorcortek.ai/}} for their help and high quality radar data capture tools used in our dataset development.



\bibliographystyle{IEEEtran}
\bibliography{IEEEabrv,IEEEtranBST/MyReference}

\begin{thebibliography}{10}
\providecommand{\url}[1]{#1}
\csname url@samestyle\endcsname
\providecommand{\newblock}{\relax}
\providecommand{\bibinfo}[2]{#2}
\providecommand{\BIBentrySTDinterwordspacing}{\spaceskip=0pt\relax}
\providecommand{\BIBentryALTinterwordstretchfactor}{4}
\providecommand{\BIBentryALTinterwordspacing}{\spaceskip=\fontdimen2\font plus
\BIBentryALTinterwordstretchfactor\fontdimen3\font minus
  \fontdimen4\font\relax}
\providecommand{\BIBforeignlanguage}[2]{{%
\expandafter\ifx\csname l@#1\endcsname\relax
\typeout{** WARNING: IEEEtran.bst: No hyphenation pattern has been}%
\typeout{** loaded for the language `#1'. Using the pattern for}%
\typeout{** the default language instead.}%
\else
\language=\csname l@#1\endcsname
\fi
#2}}
\providecommand{\BIBdecl}{\relax}
\BIBdecl

\bibitem{Ref:HighResolutionRadarDataset}
M.~{Mostajabi}, C.~M. {Wang}, D.~{Ranjan}, and G.~{Hsyu}, ``High resolution
  radar dataset for semi-supervised learning of dynamic objects,'' in
  \emph{IEEE Conference on Computer Vision and Pattern Recognition Workshops},
  2020.

\bibitem{Ref:ProbabilisticOriented}
X.~{Dong}, P.~{Wang}, P.~{Zhang}, and L.~{Liu}, ``Probabilistic oriented object
  detection in automotive radar,'' in \emph{IEEE Conference on Computer Vision
  and Pattern Recognition Workshops}, 2020, pp. 458--467.

\bibitem{Ref:DeepRadarDetector}
D.~{Brodeski}, I.~{Bilik}, and R.~{Giryes}, ``Deep radar detector,'' in
  \emph{IEEE Radar Conference}, 2019.

\bibitem{Ref:DeepOpenSpace}
F.~E. {Nowruzi}, D.~{Kolhatkar}, P.~{Kapoor}, F.~{Al Hassanat}, E.~J. {Heravi},
  R.~{Laganiere}, J.~{Rebut}, and W.~{Malik}, ``Deep open space segmentation
  using automotive radar,'' in \emph{IEEE MTT-S International Conference on
  Microwaves for Intelligent Mobility}, 2020, pp. 1--4.

\bibitem{BG:MUSIC}
R.~{Schmidt}, ``Multiple emitter location and signal parameter estimation,''
  \emph{IEEE Transactions on Antennas and Propagation}, vol.~34, no.~3, pp.
  276--280, 1986.

\bibitem{Ref:3DRadarCube}
A.~{Palffy}, J.~{Dong}, J.~F.~P. {Kooij}, and D.~M. {Gavrila}, ``Cnn based road
  user detection using the 3d radar cube,'' \emph{IEEE Robotics and Automation
  Letters}, vol.~5, no.~2, pp. 1263--1270, 2020.

\bibitem{Ref:ClusterSegmentation}
O.~{Schumann}, M.~{Hahn}, J.~{Dickmann}, and C.~{Wöhler}, ``Semantic
  segmentation on radar point clouds,'' in \emph{International Conference on
  Information Fusion}, 2018, pp. 2179--2186.

\bibitem{BG:ResNet}
K.~He, X.~Zhang, S.~Ren, and J.~Sun, ``Deep residual learning for image
  recognition,'' in \emph{IEEE conference on computer vision and pattern
  recognition}, 2016, pp. 770--778.

\bibitem{Ref:YOLOv4}
A.~{Bochkovskiy}, C.~Y. {Wang}, and H.~M. {Liao}, ``Yolov4: Optimal speed and
  accuracy of object detection,'' \emph{arXiv preprint arXiv:2004.10934}, 2020.

\bibitem{Ref:RadarCamFusion}
T.~Y. {Lim}, A.~{Ansari}, B.~{Major}, D.~{Fontijne}, M.~{Hamilton},
  R.~{Gowaikar}, and S.~{Subramanian}, ``Radar and camera early fusion for
  vehicle detection in advanced driver assistance systems,'' in \emph{Neural
  Information Processing Systems}, 2019.

\bibitem{Ref:RADVehicle}
B.~{Major}, D.~{Fontijne}, A.~{Ansari}, R.~T. {Sukhavasi}, R.~{Gowaikar},
  M.~{Hamilton}, S.~{Lee}, S.~{Grzechnik}, and S.~{Subramanian}, ``Vehicle
  detection with automotive radar using deep learning on range-azimuth-doppler
  tensors,'' in \emph{IEEE International Conference on Computer Vision
  Workshop}, 2019, pp. 924--932.

\bibitem{Ref:ClusterLSTM}
O.~{Schumann}, C.~{Wöhler}, M.~{Hahn}, and J.~{Dickmann}, ``Comparison of
  random forest and long short-term memory network performances in
  classification tasks using radar,'' in \emph{Sensor Data Fusion: Trends,
  Solutions, Applications}, 2017, pp. 1--6.

\bibitem{BG:DBSCAN}
M.~{Ester}, H.~{Kriegel}, J.~{Sander}, and X.~{Xu}, ``A density-based algorithm
  for discovering clusters in large spatial databases with noise,'' in
  \emph{International Conference on Knowledge Discovery}, 1996.

\bibitem{Ref:ClassificationRadarSpectra}
K.~{Patel}, K.~{Rambach}, T.~{Visentin}, D.~{Rusev}, M.~{Pfeiffer}, and
  B.~{Yang}, ``Deep learning-based object classification on automotive radar
  spectra,'' in \emph{IEEE Radar Conference}, 2019.

\bibitem{Ref:PointNet++}
C.~R. {Qi}, L.~{Yi}, H.~{Su}, and L.~J. {Guibas}, ``Pointnet++ deep
  hierarchical feature learning on point sets in a metric space,'' in
  \emph{Neural Information Processing Systems}, 2017, pp. 5105--5114.

\bibitem{BG:VGG}
K.~{Simonyan} and A.~{Zisserman}, ``Very deep convolutional networks for
  large-scale image recognition,'' \emph{arXiv e-prints}, 2014.

\bibitem{Ref:FasterRCNN}
S.~{Ren}, K.~{He}, R.~B. {Girshick}, and J.~{Sun}, ``Faster r-cnn: Towards
  real-time object detection with region proposal networks,'' in \emph{Neural
  Information Processing Systems}, 2015, pp. 91--99.

\bibitem{Ref:U-Net}
O.~{Ronneberger}, P.~{Fischer}, and T.~{Brox}, ``U-net: Convolutional networks
  for biomedical image segmentation,'' in \emph{Medical Image Computing and
  Computer-Assisted Intervention}, vol. 9351, 10 2015, pp. 234--241.

\bibitem{Ref:FCN}
E.~{Shelhamer}, J.~{Long}, and T.~{Darrell}, ``Fully convolutional networks for
  semantic segmentation,'' \emph{IEEE Transactions on Pattern Analysis and
  Machine Intelligence}, vol.~39, pp. 1--1, 05 2016.

\bibitem{Ref:RadarGhostObject}
J.~M. {Garcia}, R.~{Prophet}, J.~C.~F. {Michel}, R.~{Ebelt}, M.~{Vossiek}, and
  I.~{Weber}, ``Identification of ghost moving detections in automotive
  scenarios with deep learning,'' in \emph{IEEE MTT-S International Conference
  on Microwaves for Intelligent Mobility}, 2019, pp. 1--4.

\bibitem{Ref:SSD}
W.~{Liu}, D.~{Anguelov}, D.~{Erhan}, C.~{Szegedy}, S.~{Reed}, C.~{Fu}, and
  A.~{Berg}., ``Ssd: Single shot multibox detector,'' in \emph{European
  Conference on Computer Vision}.\hskip 1em plus 0.5em minus 0.4em\relax
  Springer, 2016, pp. 21--37.

\bibitem{Ref:FocalLoss}
T.~Y. {Lin}, P.~{Goyal}, R.~B. {Girshick}, K.~{He}, and P.~{Dollár}, ``Focal
  loss for dense object detection,'' in \emph{IEEE International Conference on
  Computer Vision}, 2017, pp. 2999--3007.

\bibitem{Ref:FCOS}
Z.~{Tian}, C.~{Shen}, H.~{Chen}, and T.~{He}, ``Fcos: Fully convolutional
  one-stage object detection,'' in \emph{IEEE International Conference on
  Computer Vision}, 2019, pp. 9626--9635.

\bibitem{BG:Calibration}
J.~{Peršić}, I.~{Marković}, and I.~{Petrović}, ``Extrinsic 6dof calibration
  of a radar–lidar–camera system enhanced by radar cross section estimates
  evaluation,'' \emph{Robotics and Autonomous Systems}, vol. 114, pp. 217--230,
  2019.

\bibitem{Ref:MRCNN}
K.~{He}, G.~{Gkioxari}, P.~{Dollar}, and R.~{Girshick}, ``Mask r-cnn,'' in
  \emph{IEEE International Conference on Computer Vision}, 2017, pp.
  2961--2969.

\bibitem{BG:BN}
S.~{Ioffe} and C.~{Szegedy}, ``Batch normalization: Accelerating deep network
  training by reducing internal covariate shift,'' in \emph{International
  Conference on Machine Learning}.\hskip 1em plus 0.5em minus 0.4em\relax PMLR,
  2015, pp. 448--456.

\bibitem{Ref:YOLOv1}
J.~{Redmon}, S.~{Divvala}, R.~{Girshick}, and A.~{Farhadi}, ``You only look
  once: Unified, real-time object detection,'' in \emph{IEEE conference on
  Computer Vision and Pattern Recognition}, 2016, pp. 779--788.

\bibitem{BG:CrossEntropy}
G.~E. {Nasr}, E.~A. {Badr}, and C.~{Joun}, ``Cross entropy error function in
  neural networks: Forecasting gasoline demand,'' in \emph{International
  Florida Artificial Intelligence Research Society Conference}.\hskip 1em plus
  0.5em minus 0.4em\relax Association for the Advancement of Artificial
  Intelligence, 2002, pp. 381--384.

\bibitem{Ref:SAGAN}
H.~{Zhang}, I.~J. {Goodfellow}, D.~N. {Metaxas}, and A.~{Odena},
  ``Self-attention generative adversarial networks,'' in \emph{International
  Conference on Machine Learning}.\hskip 1em plus 0.5em minus 0.4em\relax PMLR,
  2019.

\bibitem{Ref:SAUnet}
O.~{Oktay}, J.~{Schlemper}, L.~L. {Folgoc}, M.~C.~H. {Lee}, M.~P. {Heinrich},
  K.~{Misawa}, K.~{Mori}, S.~G. {McDonagh}, N.~Y. {Hammerla}, B.~{Kainz},
  B.~{Glocker}, and D.~{Rueckert}, ``Attention u-net: Learning where to look
  for the pancreas,'' \emph{arXiv preprint arXiv:1804.03999}, 2018.

\bibitem{BG:Adam}
D.~P. Kingma and J.~Ba, ``Adam: A method for stochastic optimization,''
  \emph{arXiv preprint arXiv:1412.6980}, 2014.

\end{thebibliography}
%

%
%

\end{document}